\documentclass[runningheads]{llncs}

\usepackage{latexsym}
\usepackage[utf8]{inputenc}

\usepackage{tabularx}
\usepackage{wrapfig}
\usepackage{multirow}
\usepackage{float}
\usepackage{amsmath}
\usepackage{amssymb}
\usepackage{multirow}
\usepackage{graphicx}
\usepackage{mathptmx}
\usepackage[usenames,dvipsnames]{color}
\usepackage{verbatim}
\usepackage{gensymb}
\usepackage{tabularx}
\usepackage{tabu}
\usepackage{tikz}
\usetikzlibrary{shapes}
\usetikzlibrary{plotmarks}
\usepackage{xcolor}
\usepackage{xcolor}

\def\bcbaux#1#2 #3\endbcb{%
  \colorbox{#1}{\strut#2}%
  \ifx\relax#3\relax\def\next{}\else%
    \colorbox{#1}{ \strut}%
    \allowbreak%
    \def\next{\bcbaux{#1}#3\endbcb}%
  \fi%
  \next%
}
\usepackage{hyperref}
\usepackage{url}
\usepackage{amssymb}
\usepackage{graphicx}
\usepackage{booktabs}
\usepackage{url}
\usepackage{tabu}
\usepackage{xcolor}
\usepackage{hyperref}
\newcommand{\softmaxname}[0]{\ensuremath{\textrm{softmax}}}

\newcommand{\repeatthanks}{\textsuperscript{\thefootnote}}

\title{Multi-Head Self-Attention with Role-Guided Masks}

\author{
Dongsheng Wang\thanks{Equal contribution}\and
Casper Hansen\repeatthanks\and
Lucas Chaves Lima\and
Christian Hansen\and
Maria Maistro\and
Jakob Grue Simonsen \and
Christina Lioma
}

\institute{Department of Computer Science, University of Copenhagen\\
\email{\{wang,c.hansen,lcl,chrh,mm,simonsen,c.lioma\}@di.ku.dk}}

\date{}
\authorrunning{D. Wang, C. Hansen, L. C. Lima, C. Hansen, M. Maistro, J. G. Simonsen, C. Lioma}

\begin{document}
\maketitle
\begin{abstract}

The state of the art in learning meaningful semantic representations of words is the Transformer model and its attention mechanisms. Simply put, the attention mechanisms learn to attend to specific parts of the input dispensing recurrence and convolutions. While some of the learned attention heads have been found to play linguistically interpretable roles, they can be redundant or prone to errors. We propose a method to guide the attention heads towards roles identified in prior work as important. We do this by defining role-specific masks to constrain the heads to attend to specific parts of the input, such that different heads are designed to play different roles. Experiments on text classification and machine translation using 7 different datasets show that our method outperforms competitive attention-based, CNN, and RNN baselines.

% #### OLD VERSION BELOW #######
% On one side the trained attention heads from Transformer model often play linguistically-interpretable roles, on the other, many of them can be of redundancy or even contain potential errors. Therefore, in this paper, we propose to incorporate role masks for self-attention heads, guiding them to attend to specific parts of the input, so that different heads have their own attention computation ranges and are expected to play different roles. The underlying benefits are that the guidance can not only prevent different heads from becoming duplicates, but prevent them from attending to obviously irrelevant tokens. To achieve this, we first adopt effective roles by taking full advantages of the roles that are found from the Transformers in recent studies; then, we produce the guided masks based on those roles. 

% The experiment shows a robust improvements across six classification tasks compared to the Transformer, and notably outperform all the other attention models on WMT16 English-German translation tasks. 
% % The thought-provoking point for us is that even we only employ the roles that are found from the Transformers themselves, we still gain satisfactory improvements, which indicates the existence of the potential redundancy and errors learned by Transformers. 
% Our method has the advantages of improving both the performance and the interpretability of the model. 
% The code and data involved in this study are available on github. \footnote{**github**}
\end{abstract}

\section{Introduction}
\label{sec:introduction}
The Transformer model has had great success in various tasks in Natural Language Processing (NLP). For instance, the state of the art is dominated by models such as BERT~\cite{devlin2018bert} and its extensions: RoBERTa \cite{liu2019roberta}, ALBERT \cite{lan2019albert}, SpanBERT \cite{joshi2020spanbert}, SemBERT \cite{zhang2019semantics}, and SciBERT \cite{beltagy2019scibert}, all of which are Transformer-based architectures. Due to this, recent studies have focused on developing approaches to understand how attention heads digest input texts, aiming to increase the interpretability of the model \cite{clark2019does,michel2019sixteen,voita2019analyzing}. 
% Most of these papers analyze the core of the Transformers models, the attention mechanisms. Simply put, the attention mechanisms learn to attend to some specific parts of the input dispensing recurrence and convolutions. 
The findings of those analyses are aligned: while some attention heads of the Transformer often play linguistically interpretable roles \cite{clark2019does,voita2019analyzing}, others are found to be less important and can be pruned without significantly impacting (indicating redundancy), or even improving (indicating potential errors contained in pruned heads), effectiveness \cite{michel2019sixteen,voita2019analyzing}. 

While the above studies show that the effectiveness of the attention heads is, in part, derived from different head roles, only scant prior work analyze the impact of \textit{explicitly} adopting roles for the multiple heads. Such an explicit guidance would force the heads to spread the attention on different parts of the input with the aim of reducing redundancy. This motivates the following research question: \textit{What is the impact of explicitly guiding attention heads?}
% \begin{description}
% \centering
% \item \textit{(RQ) What is the impact of explicitly \\ guiding attention heads?}
% \end{description}

To answer this question, we %put forward to 
define role-specific masks to guide the attention heads to attend to different parts of the input, such that different heads are designed to play different roles.
%such that different heads have their own attention computation ranges and are expected to play different roles. 
We first choose important roles based on findings from recent studies on interpretable Transformers roles; %detected from the Transformers in recent studies; 
then we produce masks with respect to those roles; and finally the masks are incorporated into self-attention heads to guide the attention computation. Experimental results on both text classification and machine translation on 7 different datasets show that our approach outperforms competitive attention-based, CNN, and RNN baselines.

\section{Related Work}
\label{sec:related}
The Transformer \cite{vaswani2017attention} was originally proposed as an encoder-decoder model, but has also been used successfully for transfer learning tasks, especially after being pre-trained on massive amounts of unlabeled texts. At the heart of the transformer lies the notion of multi-head self-attention, where the attention of each head is computed as:
\begin{equation}
    \label{eq:attent}
    \textrm{Attention}(Q,K,V)=\softmaxname\left(\frac{QK^T}{\sqrt{d_k}}\right) V
\end{equation}
\noindent where $Q$, is the query, $K$ is the key, $V$ is the value, and $d_k$ is the key dimension. The input to each head is a head-specific linear projection, and the Transformer uses multi-heads such that the attention for each head is concatenated for a single output.

Recently, efforts have been made to explore how the Transformer attends over different parts of the input texts \cite{clark2019does,hoover2019exbert,voita2019analyzing}.
Clark et al. \cite{clark2019does} investigate each attention head's linguistic roles, and find that particular heads refer to specific aspects of syntax. 
Voita et al. \cite{voita2019analyzing} study the importance of the different heads using layer-wise relevance propagation (LRP) \cite{ding2017visualizing}, and characterize them based on the role they perform. Furthermore, Voita et al.\ \cite{voita2019analyzing} find that not all heads are equally important and choose to prune the heads using a $L_0$ regularizer, finding that most of the non-pruned heads have specialized roles.

%In addition, \citet{voita2019analyzing} used a $L_0$ regulariser,  \cite{louizos2017learning}, to prune the least imapctful heads, and found that most of the non-pruned heads had specialised roles.

Scant prior work exists on guiding the attention heads to have a specific purpose. Strubell et al. \cite{strubell2018lingu} train the multi-head model with the first head attending to a single syntactic parent token, while the rest being regular attention heads. In contrast, we explore multiple more complex predefined roles grounded in head roles discovered in recent work. Sennrich and Haddow \cite{sennrich2016linguistic} incorporate linguistic features (e.g. sub-word tags, POS tags, etc.) as additional features into an attention encoder and decoder model for the task of machine translation, in order to enrich the model. In contrast, our method also makes use of linguistic features, but instead of enriching the input, we use these linguistic features to define the role-specific masks for guiding the attention heads.
%has the advantage of incorporating linguistic features, as defined by role-specific masks, without disturbing the language model itself.

\section{Multi-head Attention with Guided Masks}
We incorporate role-specific masks for self-attention heads, constraining them to attend to specific parts of the input. By doing this, we aim to reduce the redundancy between the heads, and force the heads to have roles identified in previous work as important. Then, we adopt a weighted gate layer to aggregate the heads.

%The primary benefit is that the role-specific masks guide the heads towards attending different parts of the input, thus aiming at reducing redundancy.

We first define the multi-head self-attention with role-specific masks in Section \ref{ss:multihead} followed by a description of each role in Section \ref{ss:gahroles}. We denote our final attention guided Transformer model as Transformer-Guided-Attn.

%; followed by the approach for generating the masks in section \ref{ss:maskgene}. 
% \lcl{Are we assuming that we can perfectly estimate the head roles of each token or is it a fact?}. 

\label{sec:attention_mechanism}

% \begin{figure}
%     \centering
%     \includegraphics[scale=0.36]{model.png}
%     \caption{The GAH architecture}
%     \label{fig:gah}
% \end{figure}

\begin{figure}
    \small
    \centering
    \includegraphics[scale=0.35]{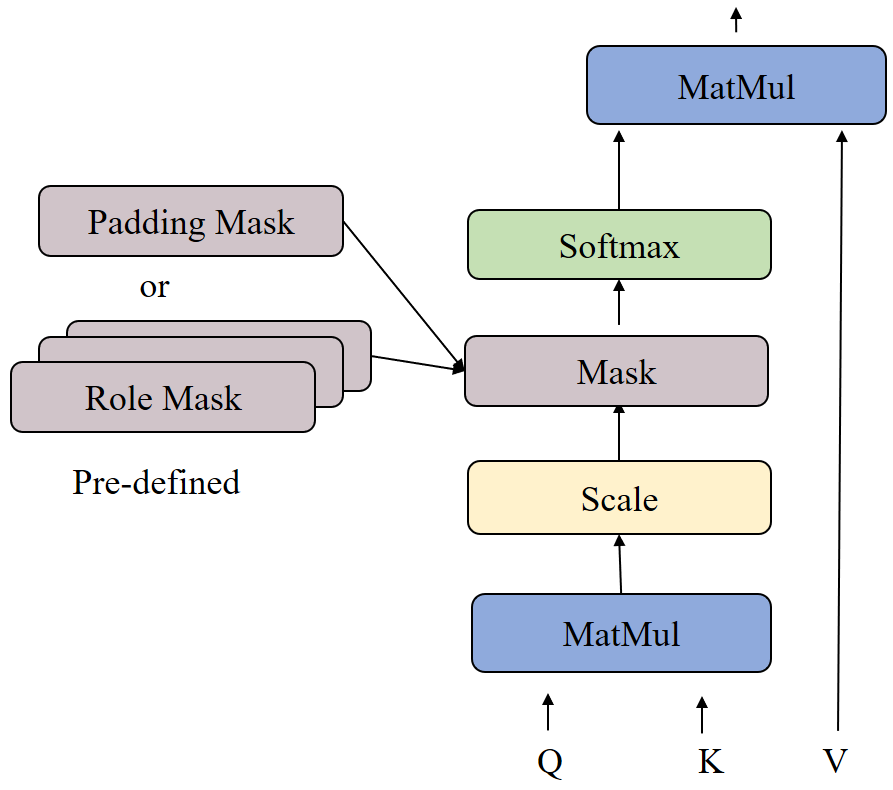}
    \vspace{-1pt}
    \caption{Scaled-dot product with role mask or padding mask.}
    \label{fig:scalemask}
    \vspace{-1pt}
\end{figure}

\subsection{Multi-head attention}
\label{ss:multihead}

We incorporate a role-specific mask into a masked attention head (mh) as:
\begin{equation}
\small
\label{eq:masked_att}
    \mathrm{mh(Q,K,V,M_r)=\softmaxname \left(\frac{QK^T+M_r}{\sqrt{d_k}}\right)V}
\end{equation}
where $M_r$ is a role-specific mask used to constrain the attention head. For an input of length $n$, $M_r$ is an $n$-by-$n$ matrix where each element is either $-\infty$ (ignore) or 0 (include). %This is different from the decoder of the Transformer that provides the option to mask some tokens, because they theoretically operate it on 1D mask by completely masking off certain tokens each time.
%
%we have $M_r = Mask(role) \cdot (-\infty)$ to disable those irrelevant positions (tokens) (by setting the values as $-\infty$). The $Mask(role)$ is an n-by-n binary matrix for each input sentence where its element $m_{ij}$ is either 1 (turn on) or 0 (turn off). This is different from the BERT or the decoder of the Transformer that provides the option to mask some tokens, because they theoretically operate it on 1D mask by completely masking off certain tokens each time.
%
For multi-head self-attention, we introduce $N$ role-specific masks for the first $N$ heads out of a total of $H$ heads ($N\leq H$). If $N$ is strictly less than $H$, then the remaining heads are regular attention heads. Based on this, the multi-head attention can be expressed as:
\begin{equation}
\small
\begin{aligned}
\label{eq:local_head}
\textrm{MultiHead}(Q,K,V) &= Concat( mh_1, mh_2, ..., mh_N, h_{N+1}, h_{N+2}, ..., h_H)W^O 
\end{aligned}
\end{equation}
where $\small{mh_i}$ is the head with a role mask, and $\small{h_i}$ is a regular head computed using Eq.~\eqref{eq:attent}.%, where $W_i^Q\in\mathbb{R}^{d_{model}X d_k}, W_i^K \in \mathbb{R}^{d_{model}Xd_k}$, $W_i^V \in \mathbb{R}^{d_{model}Xd_v}$, and $W^O \in \mathbb{R}^{hd_vXd_{model}}$ (in practice, $d_v = d_k$).

%As shown in Figure \ref{fig:scalemask}, in practice, we can default a padding mask to a regular attention head. Padding mask is for samples that have shorter length than the uniform length, with some part of the data, actually padding, should be ignored. We detail the role mask generation in section \ref{ss:maskgene}. 

A visualization of using the masks is shown in Figure~\ref{fig:scalemask}, where we associate the standard padding mask to regular attention heads. The padding masks ensure that inputs shorter than the model allowed length are padded to fit the model.

\subsection{Mask roles}
\label{ss:gahroles}
We adopt the roles detected as important by Voita et al.\ \cite{voita2019analyzing} and Clark et al. \cite{clark2019does}. We categorize them as 1) specialized (rare words and separators), 2) syntactic (dependency syntax and major relations), and 3) window (relative position) roles (see \cite{LiomaB09,Liomav08} for a linguistic basis of this categorisation). 
%The roles that they both detected are relative positions and rare words. 
%The two syntactic roles are different and complementary: major syntactic relations role from \citet{voita2019analyzing} and dependency syntax from \citet{clark2019does}. 
We include the separator role as Clark et al.~\cite{clark2019does} found that over half of BERT’s attention, in layer 6-10, focus on separators.
We describe these 5 specific roles below, which are used for creating role-specific masks. 
%To employ effective roles, we utilize roles detected from the trained Transformer models, since the trained heads can result in significant performance across NLP tasks. We detail them next.
% Therefore, we introduce specifically how we adopt the roles shown in Table \ref{tb:head_roles} in this section.

\begin{description}

\item[Rare words (RareW)] The rare words role refers to the least frequent tokens in a text. As defined by Voita et al. \cite{voita2019analyzing}, we compute IDF (inversed document frequency) scores for all tokens and use the 10 percent least frequent tokens (highest 10 percent values according to IDF) in the sentence as the target attentions.% Practically we set p equals to ten in this paper.

\item[Separator (Seprat)] The separator role guides the head to point to only separators. We extend the separator from $\{[SEP], [START], [END]\}$ to common punctuation of \{comma, semicolon, dot, question mark, exclamation point\}. 
% We assume this role can assist the text to recognize the position information as a complementary to position embedding, because the position embedding is a token-wise information while the separator indicates a multi-word (more like a clause level) position indicators.  

\item[Dependency syntax (DepSyn)] Dependency syntax role guides the head to attend to tokens with syntactic dependency relations. We assume this role can guide the head to attend to those--not adjacent--but still relevant tokens, complementary to the RelPos role (see below).

\item[Major syntactic relations (MajRel)] The major syntactic relations role guides the head to attend to the tokens involved with major syntactic relations. The four major relations defined by Voita et al. \cite{voita2019analyzing} are NSUBJ, DOBJ, AMOD, and ADVMOD. %We assume this role can capture the major semantics. 

\item[Relative Position (RelPos)] The relative position role guides the head to look at adjacent tokens, corresponding to scanning the text with a centered window of size 3.% which is to some extent similar to natural language window scanning with a fixed window size of one (or a range of three tokens). 
%In principle, the window size could be increased into more than one (e.g. two or three) as a generalization.
\end{description}

For each role, we generate the guided mask for each input sentence by first producing an $n$-by-$n$ matrix with all values as $-\infty$ (corresponding to ignoring all tokens initially). Then, we change the value of position $(i,j)$ into 0.0, referring to the query token $i$ with respect to the guided key token $j$, depending on the mask role. % by proceeding over the five roles, i.e., RareW, Seprat, DepSyn, MajRel, and RelPos.% (more details can be found in the code).

\section{Experiment}
We experimentally compare our Transformer-Guided-Attn model to competitive baselines across 7 datasets in the tasks of text classification and machine translation. 
We make the source code publicly available on GitHub\footnote{\url{https://github.com/dswang2011/guided-attention-transformer}}.

\subsection{Classification tasks}
We consider two different classification tasks: sentiment analysis and topic classification. We compare our methods against six competitive baselines: the original Transformer \cite{vaswani2017attention}; multi-scale CNNs \cite{wang2018copenhagen}; RNNs (BiLSTM) \cite{bin2016bidirectional}; directional Self-attention (DiSAN) \cite{shen2018disan} that incorporates temporal order and multi-dimensional attention into the Transformer; phrase-level self-attention (PSAN) \cite{wu2018phrase} which performs self-attention across words inside a phrase; and Transformer-Complex-Order \cite{wang2019encoding} that incorporates sequential order into the Transformer to capture ordered relationships between token positions. For the baselines implemented by us (marked in the Tables), we tune them as described in the original papers. 
For our Transformer-Guided-Attn, we consider a simple, but effective, way of selecting the combination of role-specific masks: For each layer, we fix 5 attention heads to be guided by the specific roles specified in Section \ref{ss:gahroles}, and let the remaining be regular heads. We tune the number of layers from $\{2,4,6,8\}$ and number of additional regular heads from $\{1,3\}$.

%We have the parameters of head number and layer number chosen from the set $\{2,4,6,8\}$. However, since we can have at most five masks for our method, thus, the head number for our model will be chosen as either 6 or 8 in order to accommodate 5 masks. 

\textbf{Dataset}. The statistics of the datasets are shown in Table \ref{tb:class_datasets}. We use the same splits as done by Wang et al. \cite{wang2019encoding}.

\textbf{Results}. As shown in Table \ref{tb:class_res}, we observe consistent improvements compared to the best baseline for each dataset, except on MR where we perform as well as PSAN. Compared to the original Transformer model, we obtain accuracy gains of up to 2.96\%, depending on the dataset, thus showing a notable performance impact from guiding the attention heads. Compared to DiSAN and PSAN, our proposed Transformer-Guided-Attn obtains consistent improvements over the original Transformer across all datasets, while DiSAN and PSAN both have lower performance for TREC and SUBJ.

%where both of these perform worse than the original transformer for TREC and SUBJ.
%we can observe consistent improvements compared with the baselines, with N masks, N-1 masks or only 1 mask. We can also observe that our method achieve the best or the same best with the baseline and the other methods. 

\begin{table}[t]
\centering
\makebox[0pt][c]{\parbox{1.0\textwidth}{%
    \begin{minipage}[b]{0.49\hsize}\centering
\caption{\label{tb:class_datasets}Classification dataset statistics. CV means 10-fold cross validation. 
%The splits for TREC and SST are consistent with in \cite{wang2019encoding}
}
\centering
\resizebox{.99\textwidth}{!}{
\begin{tabular}{@{}lllllll@{}}
\toprule
Dataset & Train & Test & Task & Vocab. & Class \\ \midrule
CR& 4k & CV & Product review & 6k & 2 \\
TREC & 5.4k & 0.5k & Question & 10k & 6 \\
SUBJ & 10k & CV & Subjectivity & 21k & 2 \\
MPQA & 11k & CV & Opinion polarity & 6k & 2 \\
MR  & 11.9k & CV & Movie review & 20k & 2 \\
SST & 67k & 2.2k & Movie review & 18k & 2 \\
 \bottomrule
\end{tabular}}
    \end{minipage}
    \hfill
    \begin{minipage}[b]{0.49\hsize}\centering
\caption{\label{tb:trans_Res} Machine translation results. $\star$ marks scores reported from other papers.}
\resizebox{.85\textwidth}{!}{
\begin{tabular}{@{}llc@{}}
\toprule
Method & BLEU &  \\ \midrule
Transformer \cite{vaswani2017attention} & 34.3 &  \\
AED + Linguistic \cite{sennrich2016linguistic} $\star$ & 28.4 &  \\
AED + BPE \cite{sennrich2016edinburgh} $\star$ & 34.2 &  \\
Tensorized Transformer \cite{ma2019tensorized} $\star$ & 34.9 &  \\
Transformer-Complex-Order \cite{wang2019encoding} $\star$ & 35.8 &  \\
\hline
Transformer-Guided-Attn (ours) & \textbf{38.8} &  \\
%Transformer-Guided-Attn (N-1) & 38.42 &  \\ 
%Transformer-Guided-Attn (1) & 37.98 &  \\
\bottomrule
\end{tabular}}
    \end{minipage}
}}
% \vspace{-10pt}
\end{table}

\begin{table}[t]
%\small
\centering
\caption{\label{tb:class_res} Classification results (accuracy \%). $\star$ marks scores reported from other papers.
}
\resizebox{.9\textwidth}{!}{
\begin{tabular}{lcccccc} % lllllll
\toprule
Method & CR & TREC & SUBJ & MPAQ & MR & SST \\ \midrule
Transformer \cite{vaswani2017attention} & 82.0 & 91.8 & 93.2 & 88.6 & 77.7 & 81.8 \\
Multi-scale CNNs \cite{wang2018copenhagen} & 81.2 & 93.1 & 93.3 & 89.1 & 77.8 & 80.9 \\
BiLSTM \cite{bin2016bidirectional}  & 82.6 & 92.4 & 93.6 & 88.9 & 78.4 & 81.1 \\
DiSAN (Directional Self-Attention) \cite{shen2018disan}$\star$ & 84.1 & 88.3 & 92.2 & 89.5 & 79.7 & 82.9 \\
PSAN (phrase-level Self-Attention) \cite{wu2018phrase}$\star$ & 84.2 & 89.1 & 91.9 & 89.9 & \textbf{80.0} & 83.8 \\
Transformer-Complex-Order \cite{wang2019encoding}$\star$ & 80.6 & 89.6 & 89.5 & 86.3 & 74.6 & 81.3 \\
\hline
%Transformer-Guided-Attn (N) & \textbf{84.4} & 92.2 & \textbf{93.8} & 90.0 & 79.9 & \textbf{84.2} \\
%Transformer-Guided-Attn (N-1) & 82.6 & \textbf{93.6} & 93.20 & \textbf{90.7} & \textbf{80.0} & 83.6 \\
%Transformer-Guided-Attn (1) & 82.3 & 93.2 & 93.19 & 89.91 & 79.76 & 82.8 \\ \bottomrule
Transformer-Guided-Attn (ours) & \textbf{84.4} & \textbf{93.6} & \textbf{93.8} & \textbf{90.7} & \textbf{80.0} & \textbf{84.2} \\ \bottomrule
\end{tabular}}
% \vspace{-10pt}
\end{table}

\subsection{Translation Task}
We use the standard WMT 2016 English-German dataset \cite{sennrich2016edinburgh} and use four baselines: Attentional encoder-decoder (AED)  \cite{sennrich2016linguistic} with linguistic features including morphological, part-of-speech, and syntactic dependency labels as additional embedding space; AED with Byte-pair encoding (BPE) \cite{sennrich2016edinburgh} subword segmentation for open-vocabulary translation; the tensorized Transformer \cite{ma2019tensorized}; and the Transformer-Complex-order \cite{wang2019encoding}. The first two models are extensions on top of the basic AED \cite{bahdanau2014neural}. For the models we implement, we follow the same tuning as in the classification experiments.
%We trained 6 heads and 2 layers for Transformer-Guided-Attn (N), 8 heads and 4 layers for Transformer-Guided-Attn (N-1), and 6 heads and 6 layers for the Transformer-Guided-Attn (1) (BothDi mask). 
We evaluate the machine translation performance using the Bilingual Evaluation Understudy (BLEU) measure.

\textbf{Results}. Our Transformer-Guided-Attn consistently outperforms the competitive baselines. Specifically, we observe gains of 8.2\% compared to the best baseline, Trans\-former-Complex-Order, and close to 13\% compared to the original Transformer. These gains are even larger than the results for the classification experiments, thus highlighting a significant performance impact from guiding the attention heads for the task of machine translation.
\subsection{Ablation study}
\begin{figure}[t]
\centering
\includegraphics[width=0.7\linewidth]{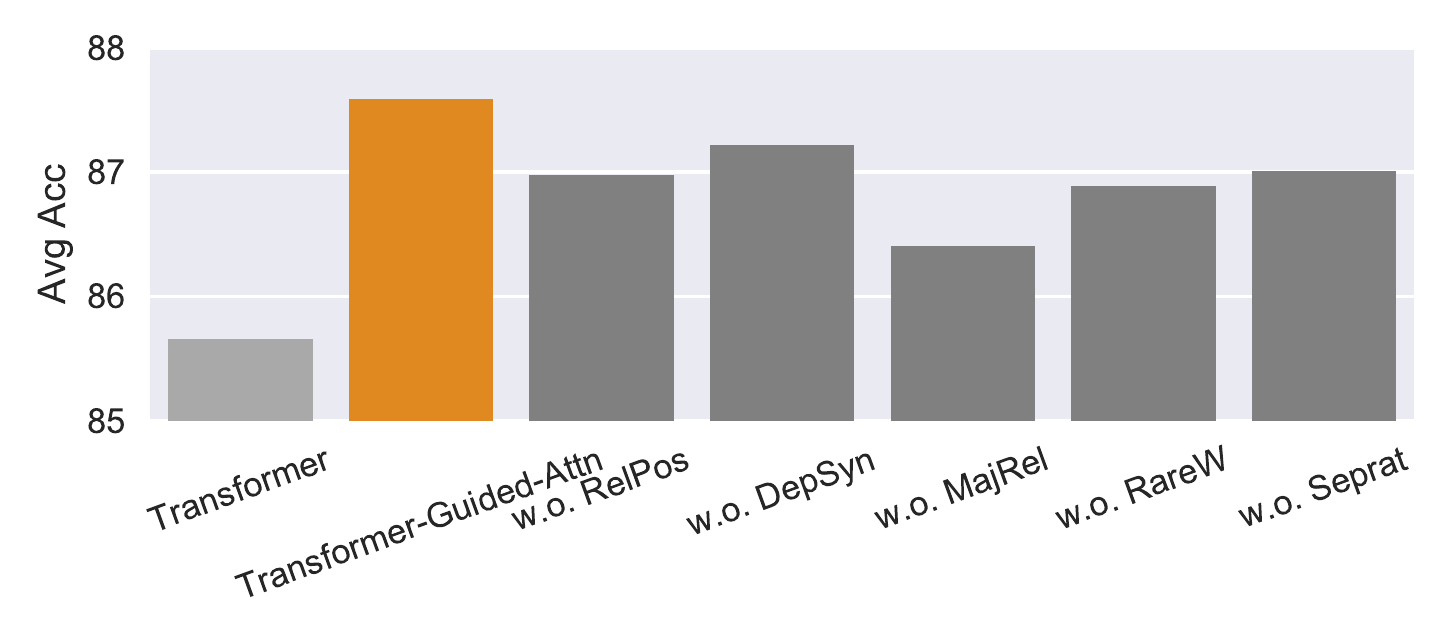}
\vspace{-15pt}
\caption{\label{fig:abl_min1} Ablation study of Transformer-Guided-Attn when dropping each role individually.}
%\label{fig:abl_head}
\vspace{-1pt}
\end{figure}

We now consider the performance impact associated with each role-specific mask. For each classification dataset, we run configurations of our Transformer-Guided-Attn with each role-specific mask excluded once and replaced with a default padding mask used in the Transformer. The average accuracy drop associated with excluding each role-specific mask is shown in Figure \ref{fig:abl_min1}, which also includes the average accuracy of the Transformer and our Transformer-Guided-Attn using all role-specific masks.
We observe that the removal of each role has a negative impact on performance, where the major syntactic relations role (MajRel) has the largest impact. Thus, collectively all roles contribute to the performance of the full Transformer-Guided-Attn model.

\section{Conclusion}
We presented Transformer-Guided-Attn, a method to explicitly guide the attention heads of the Transformer using role-specific masks. The motivation of this explicit guidance is to force the heads to spread their attention on different parts of the input with the aim of reducing redundancy among the heads. Our experiments demonstrated that incorporating multiple role masks into multi-head attention can consistently improve performance on both classification and machine translation tasks.
%The resulting multi-head model has the great advantages of improving both the performance and the interpretability of the model.

As future work, we plan to explore additional roles for masking, as well as evaluating the impact of including it for pre-training language representation models such as BERT \cite{devlin2018bert}.

%In the future, we plan to evaluate the pre-training power of our guided attention model by generating a contextual representation on a large corpus.

\section{Acknowledgments}
% This work is supported by the Quantum Access and Retrieval Theory (QUARTZ) project, which has received funding from the European Union’s Horizon 2020 research and innovation programme under the Marie Sklodowska-Curie grant agreement No. 721321

This work is supported by the European Union's Horizon 2020 research and innovation programme under the Marie Sk\l{}odowska-Curie grant agreement No.~721321 (QUARTZ project) and No.~893667 (METER project).

\clearpage

%\bibliographystyle{acl_natbib}
%\bibliography{anthology,emnlp2020}
\bibliographystyle{splncs04}
\bibliography{emnlp2020}

\begin{thebibliography}{10}
\providecommand{\url}[1]{\texttt{#1}}
\providecommand{\urlprefix}{URL }
\providecommand{\doi}[1]{https://doi.org/#1}

\bibitem{bahdanau2014neural}
Bahdanau, D., Cho, K., Bengio, Y.: Neural machine translation by jointly
  learning to align and translate. In: Bengio, Y., LeCun, Y. (eds.) 3rd
  International Conference on Learning Representations, {ICLR} 2015, San Diego,
  CA, USA, May 7-9, 2015, Conference Track Proceedings (2015),
  \url{http://arxiv.org/abs/1409.0473}

\bibitem{beltagy2019scibert}
Beltagy, I., Cohan, A., Lo, K.: Scibert: Pretrained contextualized embeddings
  for scientific text. CoRR  \textbf{abs/1903.10676} (2019),
  \url{http://arxiv.org/abs/1903.10676}

\bibitem{bin2016bidirectional}
Bin, Y., Yang, Y., Shen, F., Xu, X., Shen, H.T.: Bidirectional long-short term
  memory for video description. In: Proceedings of the 24th ACM international
  conference on Multimedia. pp. 436--440 (2016)

\bibitem{clark2019does}
Clark, K., Khandelwal, U., Levy, O., Manning, C.D.: What does {BERT} look at?
  an analysis of {BERT}'s attention. arXiv preprint arXiv:1906.04341  (2019)

\bibitem{devlin2018bert}
Devlin, J., Chang, M., Lee, K., Toutanova, K.: {BERT:} pre-training of deep
  bidirectional transformers for language understanding. In: Burstein, J.,
  Doran, C., Solorio, T. (eds.) Proceedings of the 2019 Conference of the North
  American Chapter of the Association for Computational Linguistics: Human
  Language Technologies, {NAACL-HLT} 2019, Minneapolis, MN, USA, June 2-7,
  2019, Volume 1. pp. 4171--4186. Association for Computational Linguistics
  (2019). \doi{10.18653/v1/n19-1423},
  \url{https://doi.org/10.18653/v1/n19-1423}

\bibitem{ding2017visualizing}
Ding, Y., Liu, Y., Luan, H., Sun, M.: Visualizing and understanding neural
  machine translation. In: Proceedings of the 55th Annual Meeting of the
  Association for Computational Linguistics (Volume 1: Long Papers). pp.
  1150--1159 (2017)

\bibitem{hoover2019exbert}
Hoover, B., Strobelt, H., Gehrmann, S.: exbert: A visual analysis tool to
  explore learned representations in transformers models. arXiv preprint
  arXiv:1910.05276  (2019)

\bibitem{joshi2020spanbert}
Joshi, M., Chen, D., Liu, Y., Weld, D.S., Zettlemoyer, L., Levy, O.: Spanbert:
  Improving pre-training by representing and predicting spans. Transactions of
  the Association for Computational Linguistics  \textbf{8},  64--77 (2020)

\bibitem{lan2019albert}
Lan, Z., Chen, M., Goodman, S., Gimpel, K., Sharma, P., Soricut, R.: Albert: A
  lite {BERT} for self-supervised learning of language representations. In:
  International Conference on Learning Representations (2020)

\bibitem{LiomaB09}
Lioma, C., Blanco, R.: Part of speech based term weighting for information
  retrieval. In: Boughanem, M., Berrut, C., Mothe, J., Soul{\'{e}}{-}Dupuy, C.
  (eds.) Advances in Information Retrieval, 31th European Conference on {IR}
  Research, {ECIR} 2009, Toulouse, France, April 6-9, 2009. Proceedings.
  Lecture Notes in Computer Science, vol.~5478, pp. 412--423. Springer (2009).
  \doi{10.1007/978-3-642-00958-7\_37},
  \url{https://doi.org/10.1007/978-3-642-00958-7\_37}

\bibitem{Liomav08}
Lioma, C., van Rijsbergen, C.J.K.: Part of speech n-grams and information
  retrieval. French Review of Applied Linguistics, Special issue on Information
  Extraction and Linguistics  \textbf{XIII}(2008/1),  9--22 (2008),
  \url{https://www.cairn-int.info/article-E_RFLA_131_0009--part-of-speech-n-grams-and-information.htm}

\bibitem{liu2019roberta}
Liu, Y., Ott, M., Goyal, N., Du, J., Joshi, M., Chen, D., Levy, O., Lewis, M.,
  Zettlemoyer, L., Stoyanov, V.: Roberta: A robustly optimized bert pretraining
  approach. arXiv preprint arXiv:1907.11692  (2019)

\bibitem{ma2019tensorized}
Ma, X., Zhang, P., Zhang, S., Duan, N., Hou, Y., Zhou, M., Song, D.: A
  tensorized transformer for language modeling. In: Advances in Neural
  Information Processing Systems. pp. 2229--2239 (2019)

\bibitem{michel2019sixteen}
Michel, P., Levy, O., Neubig, G.: Are sixteen heads really better than one? In:
  Advances in Neural Information Processing Systems. pp. 14014--14024 (2019)

\bibitem{sennrich2016linguistic}
Sennrich, R., Haddow, B.: Linguistic input features improve neural machine
  translation. In: Proceedings of the First Conference on Machine Translation,
  {WMT} 2016, colocated with {ACL} 2016, August 11-12, Berlin, Germany. pp.
  83--91. The Association for Computer Linguistics (2016).
  \doi{10.18653/v1/w16-2209}, \url{https://doi.org/10.18653/v1/w16-2209}

\bibitem{sennrich2016edinburgh}
Sennrich, R., Haddow, B., Birch, A.: {E}dinburgh neural machine translation
  systems for {WMT} 16. In: Proceedings of the First Conference on Machine
  Translation: Volume 2, Shared Task Papers. pp. 371--376. Association for
  Computational Linguistics, Berlin, Germany (Aug 2016).
  \doi{10.18653/v1/W16-2323}, \url{https://www.aclweb.org/anthology/W16-2323}

\bibitem{shen2018disan}
Shen, T., Zhou, T., Long, G., Jiang, J., Pan, S., Zhang, C.: Disan: Directional
  self-attention network for rnn/cnn-free language understanding. In:
  Thirty-Second AAAI Conference on Artificial Intelligence (2018)

\bibitem{strubell2018lingu}
Strubell, E., Verga, P., Andor, D., Weiss, D., McCallum, A.:
  Linguistically-informed self-attention for semantic role labeling. In:
  Proceedings of the 2018 Conference on Empirical Methods in Natural Language
  Processing. pp. 5027--5038. Association for Computational Linguistics,
  Brussels, Belgium (Oct-Nov 2018). \doi{10.18653/v1/D18-1548},
  \url{https://www.aclweb.org/anthology/D18-1548}

\bibitem{vaswani2017attention}
Vaswani, A., Shazeer, N., Parmar, N., Uszkoreit, J., Jones, L., Gomez, A.N.,
  Kaiser, {\L}., Polosukhin, I.: Attention is all you need. In: Advances in
  neural information processing systems. pp. 5998--6008 (2017)

\bibitem{voita2019analyzing}
Voita, E., Talbot, D., Moiseev, F., Sennrich, R., Titov, I.: Analyzing
  multi-head self-attention: Specialized heads do the heavy lifting, the rest
  can be pruned. In: Korhonen, A., Traum, D.R., M{\`{a}}rquez, L. (eds.)
  Proceedings of the 57th Conference of the Association for Computational
  Linguistics, {ACL} 2019, Florence, Italy, July 28- August 2, 2019, Volume 1:
  Long Papers. pp. 5797--5808. Association for Computational Linguistics
  (2019). \doi{10.18653/v1/p19-1580},
  \url{https://doi.org/10.18653/v1/p19-1580}

\bibitem{wang2019encoding}
Wang, B., Zhao, D., Lioma, C., Li, Q., Zhang, P., Simonsen, J.G.: Encoding word
  order in complex embeddings. In: 8th International Conference on Learning
  Representations, {ICLR} 2020, Addis Ababa, Ethiopia, April 26-30, 2020.
  OpenReview.net (2020), \url{https://openreview.net/forum?id=Hke-WTVtwr}

\bibitem{wang2018copenhagen}
Wang, D., Simonsen, J.G., Larsen, B., Lioma, C.: The {C}openhagen team
  participation in the factuality task of the competition of automatic
  identification and verification of claims in political debates of the
  clef-2018 fact checking lab. CLEF (Working Notes)  \textbf{2125} (2018)

\bibitem{wu2018phrase}
Wu, W., Wang, H., Liu, T., Ma, S.: Phrase-level self-attention networks for
  universal sentence encoding. In: Proceedings of the 2018 Conference on
  Empirical Methods in Natural Language Processing. pp. 3729--3738 (2018)

\bibitem{zhang2019semantics}
Zhang, Z., Wu, Y., Zhao, H., Li, Z., Zhang, S., Zhou, X., Zhou, X.:
  Semantics-aware {BERT} for language understanding. CoRR
  \textbf{abs/1909.02209} (2019), \url{http://arxiv.org/abs/1909.02209}

\end{thebibliography}

\end{document}